\documentclass[sigconf]{acmart}

\usepackage{booktabs} 
\usepackage{multirow}
\usepackage{amsmath}
\usepackage[noend]{algpseudocode}
\usepackage{graphicx}
\usepackage{hhline}
\usepackage{url}
\usepackage{amsmath,amssymb}
\usepackage{subcaption}

\usepackage{bm}
\usepackage{mathtools}
\newcommand{\bs}[1]{\ensuremath{\bm{\mathit{#1}}}}

\newcommand{\squishlist}{
 \begin{list}{$\bullet$}
  {  \setlength{\itemsep}{0pt}
     \setlength{\parsep}{3pt}
     \setlength{\topsep}{3pt}
     \setlength{\partopsep}{0pt}
     \setlength{\leftmargin}{2em}
     \setlength{\labelwidth}{1.5em}
     \setlength{\labelsep}{0.5em}
} }
\newcommand{\squishlisttight}{
 \begin{list}{$\bullet$}
  { \setlength{\itemsep}{0pt}
    \setlength{\parsep}{0pt}
    \setlength{\topsep}{0pt}
    \setlength{\partopsep}{0pt}
    \setlength{\leftmargin}{2em}
    \setlength{\labelwidth}{1.5em}
    \setlength{\labelsep}{0.5em}
} }

\newcommand{\squishdesc}{
 \begin{list}{}
  {  \setlength{\itemsep}{0pt}
     \setlength{\parsep}{3pt}
     \setlength{\topsep}{3pt}
     \setlength{\partopsep}{0pt}
     \setlength{\leftmargin}{1em}
     \setlength{\labelwidth}{1.5em}
     \setlength{\labelsep}{0.5em}
} }

\newcommand{\squishend}{
  \end{list}
}

\makeatletter
\providecommand*{\diff}%
        {\@ifnextchar^{\DIfF}{\DIfF^{}}}
\def\DIfF^#1{%
        \mathop{\mathrm{\mathstrut d}}%
                \nolimits^{#1}\gobblespace
}
\def\gobblespace{%
        \futurelet\diffarg\opspace}
\def\opspace{%
        \let\DiffSpace\!%
        \ifx\diffarg(%
                \let\DiffSpace\relax
        \else
                \ifx\diffarg\[%
                        \let\DiffSpace\relax
                \else
                        \ifx\diffarg\{%
                                \let\DiffSpace\relax
                        \fi\fi\fi\DiffSpace}

\numberwithin{equation}{section}

\renewcommand\footnotetextcopyrightpermission[1]{} 
\title{Sequential Variational Autoencoders for Collaborative Filtering}

\author{Noveen Sachdeva} 
\authornote{Work done while in internship at ICAR-CNR.}
\affiliation{   
	\institution{International Institute of Information Technology}
	\city{Hyderabad, India} 
} 
\email{noveen.sachdeva@research.iiit.ac.in}

 \author{Giuseppe Manco}
 \orcid{0000-0001-9672-3833}
 \affiliation{%
   \institution{ICAR-CNR}
   \streetaddress{Via Bucci, 8/9c}
   \city{Rende}
   \country{Italy}
   \postcode{87036}
 }
 \email{giuseppe.manco@icar.cnr.it}

 \author{Ettore Ritacco}
 \affiliation{%
   \institution{ICAR-CNR}
   \streetaddress{Via Bucci, 8/9c}
   \city{Rende}
   \state{Italy}
   \postcode{87036}
 }
 \email{ettore.ritacco@icar.cnr.it}

\author{Vikram Pudi}
\affiliation{   
	\institution{International Institute of Information Technology}
	\city{Hyderabad, India}  
} 
\email{vikram@iiit.ac.in}

\begin{abstract}
Variational autoencoders were proven successful in domains
such as computer vision and speech processing. Their adoption for
modeling user preferences is still unexplored, although recently it
is starting to gain attention in the current literature.  
In this work, we propose a model which extends variational
autoencoders by exploiting the rich information present in the past
preference history. We introduce a recurrent version of the VAE,
where instead of passing a subset of the whole history regardless of
temporal dependencies, we rather pass the consumption 
sequence subset through a recurrent neural network. At each
time-step of the RNN, the sequence is fed through a series of
fully-connected layers, the output of which models the probability
distribution of the most likely future preferences. We show that
handling temporal information is crucial for improving the accuracy of
the VAE: In fact, our model beats the current
state-of-the-art by valuable margins because of its ability to capture
temporal dependencies among the user-consumption sequence using the
recurrent encoder still keeping the fundamentals of
 variational autoencoders intact.
\end{abstract}



\begin{document}

\maketitle

\section{Introduction}
The growing diffusion of Web-based services allows an
increasingly large number of users to access and consume online
content. People access web pages, 
purchase items in e-commerce sites, watch online content through
streaming services, or interact within social networks and media. 
Understanding the factors that characterize user preferences
and shape their future behavior is crucial in order to provide
users with a better experience through the recommendation of new
content and data that users are likely to appreciate.

Collaborative filtering approaches to recommendation were extensively
investigated by the current literature
\cite{Aggarwal:2016:RST:2931100}. Among these, latent variable models   
\cite{Hofmann:2004,Ruslan:2008,Wang:2011,Kabbur:2013}
gained substantial attention, due to their capabilities in modeling
the hidden causal relationships that ultimately influence user
preferences.  Recently, however, new approaches based on neural
architectures (see \cite{ZhangYS17aa} for a comprehensive survey) were
proposed, achieving competitive performance with respect to the
current state of the art. Also, new paradigms based on the combination
of deep learning and
probabilistic latent variable modeling 
\cite{kingma14,RezendeMW14} were proven quite
successful in domains such as computer vision and speech
processing. However, their adoption for modeling user preferences is
still unexplored, although recently it is starting to gain attention
\cite{Liang:2018:VAC:3178876.3186150,Li:2017}.

The aforementioned approaches rely on the ``bag-of-word''
assumption: when considering a user and her preferences, the order of
such preferences can be neglected and all preferences are
exchangeable. This assumption works with general user trends which
reflect a long-term behavior. However, it fails to capture the
short-term preferences that are specific of several application
scenarios, especially in the context of the Web. Sequential
data can express causalities and dependencies
that require ad-hoc modeling and algorithms. And in fact, efforts to
capture this notion of causality have been made, both in the context of latent variable modeling
\cite{Rendle:2010,Barbieri:2013,Tavakol:2014,He:2017} and deep learning
\cite{Hidasi:2016,Devooght:2017,Wu:2017,Tang:2018}.  

In this paper, we consider the task of sequence recommendation from the
perspective of combining deep learning and latent variable
modeling. Inspired by the approach in
\cite{Liang:2018:VAC:3178876.3186150}, we assume that at a given
timestamp the choice of a given item is influenced by a latent
factor that models user trends and preferences. However, the latent
factor itself can be influenced by user history and modeled to capture
both long-term preferences and short-term behavior. As a matter of
fact, the recent studies in the recurrent neural network literature
\cite{ChungGCB14,Greff:2017,Vaswani:2017} demonstrate the capabilities of these
architectures in capturing both long-term and short-term
relationships. It is hence natural to exploit them in a variational
setting.  

Our contribution can be summarized as follows. 
\begin{itemize}
\item We review the framework of Variational Autoencoders and discuss
  its adoption for the problem of modeling implicit preferences in a
  collaborative filtering scenario. 
\item We extend the framework to the case of sequential
  recommendation, where user's preferences exhibit temporal
  dependencies. In particular, we discuss different modeling
  alternatives for tackling this task through the adoption of
  recurrent neural network architectures that model latent 
  dependencies at different abstraction levels. 
\item We evaluate the proposed framework on standard benchmark
  datasets, by showing that \emph{(a)} approaches not considering
  temporal dynamics are not totally adequate to model user
  preferences, and \emph{(b)} the combination of latent variable
  and temporal dependency modeling produces a substantial
  gain, even with regard to other approaches that only focus on
  temporal dependencies through recurrent relationships. 
\end{itemize}

The rest of the paper is organized as follows. Section
\ref{sec:related} discusses the recent contributions in the current
literature and provides a systematic review of the approaches related
to our task of interest. Sections~\ref{sec:back} and~\ref{sec:vae} propose
the modeling of user preferences in a variational setting, by
illustrating how the general framework can be adapted to the case of
temporally ordered dependencies. The effectiveness of the proposed
modeling is illustrated in section~\ref{sec:exp}, and pointers to
future developments are discussed in section~\ref{sec:conc}.

\section{Related Work}
\label{sec:related}

Recommender systems have been extensively studied over the last two
decades. Within the collaborative filtering framework, recommendation is
essentially modeled as a prediction problem: Given a user and an item,
we would like to predict the preference of the user for that item,
exploiting the user's past choices, i.e. her history.

Most approaches disregard the temporal
order of the preferences in a user's history. Among these,
latent variable models
\cite{Hofmann:2004,Ruslan:2008,Salakhutdinov:2008,Rendle:2009,flda,Ning:2011,Wang:2011,Rendle:2012,Kabbur:2013,2014Barbieri,BM2011}
were proven extremely effective
in modeling user preferences and providing reliable recommendations.    
Essentially, these approaches embed users and items into latent spaces
that translate relatedness into geometrical closeness. The latent
embeddings can be used to decompose
the large sparse \emph{preference matrix} 
\cite{Salakhutdinov:2008,Ruslan:2008,flda}, to devise item
similarity~\cite{Ning:2011, Kabbur:2013}, or more generally to
parameterize probability distributions for item preference \cite{Rendle:2009,
  Hofmann:2004,Wang:2011,Rendle:2012} and sharpen the prediction
quality by means of meaningful priors.

The recent literature is currently focusing on deep learning, which
shows substantial advantages over traditional approaches. 
For example, \emph{Neural Collaborative Filtering} (NCF) \cite{He17}
generalizes matrix factorization to a non-linear setting, where users,
items and preferences are modeled through a simple multilayer
perceptron network that exploits latent factor transformations.

Notably, prominent deep learning approaches to collaborative
filtering are based on the idea of autoencoding the features from the
preference matrix.  \emph{AutoRec} \cite{Sedhain15} exploits
autoencoders to encode preference histories. Unseen preferences can
be devised by looking at the reconstructed decoding, which is shaped to
include scores for all possible items of interest.
Autoencoders are also amenable to consider side information
\cite{strub15} to mitigate the sparsity of the data and to tackle the
cold start problem.

Hybrid approaches that integrate latent variable modeling and deep
learning have also gained attention. \emph{Collaborative Deep
  Learning} \cite{Wang15} embeds a \emph{stacked denoising
  autoencoder} \cite{Vincent10} into a Bayesian matrix factorization
setting.  Similarly, \emph{AutoSVD++} \cite{Zhang17} exploits
the notion of \emph{contractive autoencoder} \cite{Rifai:2011} to
learn latent item representations that are integrated into the
\emph{SVD++} model \cite{Koren08}.

The introduction of the variational autoencoding framework
\cite{kingma14,RezendeMW14} has suggested a tighter coupling
between deep learning and latent variable modeling.
\emph{Collaborative Variational Autoencoder} (CVA) \cite{Li:2017} and
\emph{Hybrid Variational Autoencoder} (HVAE)
\cite{2018arXiv180801006G} exploit side information to feed a
variational autoencoder whose goal is to produce a latent
representation of the items. In CVA, the preference matrix is hence
modeled by combining user and item embeddings with the item
latent representations, while HVAE uses another variational
autoencoder to reproduce the whole users' preference history. 
By contrast, 
\cite{Liang:2018:VAC:3178876.3186150} proposes a neural generative
model where a user's history is modeled through a multinomial
likelihood conditioned to a latent user representation which in turn
is modeled through a variational autoencoder.


Within the context of collaborative filtering, a strong effort has
also been made to model temporal dynamics within the history of user preferences
\cite{Quadrana:2018}. 
The \emph{Factorizing Personalized Markov Chain} model (FPMC)
\cite{Rendle:2010}, for example, proposes a combination of matrix
factorization and Markov chains. FPMC considers personalized
first-order transition probabilities between items, which are modeled
by decomposing the underlying tensor through user and item
embeddings.
Transition probabilities can also be measured by exploiting more
sophisticated modeling \cite{HeM16,He:2017}, where users are mapped into 
translation (latent) vectors operating on item sequences and
consequently a transition corresponds to a geometric affinity of these
latent vectors. 
Orthogonally, Markov dependencies can also be exploited to model
dependencies between latent variables~\cite{Barbieri:2013}, thus
resulting in richer formalizations and more accurate recommendations. 

Recently, a revamped interest for sequence-based recommendation has
taken place, motivated by both the success of recurrent neural
networks \cite{Cho14,Greff:2017} in domains such as language modeling,
and the need to focus on \emph{session-based} 
recommendations~\cite{Hidasi:2016, Tan16}, i.e. recommendation that do not rely 
on a user model and instead can cope with single anonymous preference
sessions. 

\emph{GRU4Rec} \cite{Hidasi:2016} proposes a recurrent neural network
model based on \emph{Gated Recurrent Units} to predict the next item
in a user session, based on the history seen so far. Since its
introduction, this model has witnessed several evolutions, and similar
architectures were proposed in
\cite{Devooght:2017,Liu:2016,Twardowski:2016, Wu16,Tan16,Jannach:2017,Quadrana:2017}.
Recurrent networks were also exploited to strengthen matrix
factorization, by producing history-aware embeddings of users and
items. The \emph{RRN} approach proposed in \cite{Wu:2017} combines two
recurrent networks, whose output at any time-step, relative to a specific
user and item, can be hence exploited to predict the current
preference.

Finally, the CASER (\emph{Convolutional Sequence
  Embedding Recommendation}) model \cite{Tang:2018} proposes an
approach that departs from RNN modeling and instead exploits a
\emph{convolutional neural network}, by transforming a sequence into
a matrix built from the concatenation of the embeddings of the items
appearing in the sequence. The matrix can hence feed convolutional
layers that can extract relevant useful features for predicting
the next items.

\section{Background}\label{sec:back}

The framework of variational autoencoders draws from the idea of
latent variable modeling \cite{Murphy:2012:MLP:2380985}. Essentially,
we can assume a $K$-dimensional 
latent variable space $\mathcal{Z}$, upon which we can devise a probability
density function $P_{\theta}(\bs{z})$ for
$\bs{z}\in\mathcal{Z}$, where $\theta$ represents a set of density
parameters. The datapoints
$\bs{X} = \left\{\bs{x}_{1}, \ldots, \bs{x}_{M}\right\}$ we observe can be
modeled through a dependency $P_\phi(\bs{x}|\bs{z})$, so 
that the overall likelihood of $\bs{X}$ can be specified by marginalizing
over $\mathcal{Z}$: 
\begin{align*}
P(\bs{X}) = & \prod_i P(\bs{x}_i) = \prod_i\int P_\phi\left(\bs{x}_i|\bs{z}\right) P_\theta(\bs{z}) \diff \bs{z} \, .
\end{align*}

Within the classical VAE framework \cite{kingma14,RezendeMW14}, \bs{z} is assumed to
be distributed according to a standard normal distribution, that is
$\theta = \{\mathbf{0},\mathbf{I}_K\}$ and consequently
$\mathbf{z} \sim \mathcal{N}(\mathbf{0},\mathbf{I}_K)$. The 
key intuition here is that even complex dependencies can be generated
starting from normally distributed variables. Thus, $P_\phi(\bs{x}_i|\bs{z})$ is
parameterized by the function $f_\phi(\bs{z})$, representing any
(unknown) transformation of $z$ that expresses such a dependency. We
can model this transformation through a neural network, so that
$\phi$ represents the set of network parameters to be optimized.
Hence, we can model the inference problem for $P(\bs{X})$ as
optimization problem, where we aim at finding the optimal parameter
set $\phi$ that maximize $P(\bs{X})$. 

However, $P(\bs{X})$ is typically intractable and
we need to find an approximation. Variational inference
\cite{Blei2017} usually
tackles this approximation by introducing a proposal distribution
$Q(\bs{z}|\bs{x})$, which approximates the true posterior
$P(\bs{z}|\bs{x})$. The relationship between the likelihood and the
proposal distribution is given by the following equation: 
\begin{align*}
  \log P(\mathbf{x}) & = \int Q(\bs{z}|\bs{x}) \log P(\bs{x}) \diff \bs{z}\\
  & =  \int Q(\bs{z}|\bs{x}) \log \frac{P(\bs{x}|\bs{z})P(\bs{z})}{P(\bs{z}|\bs{x})} \diff \bs{z}\\
  & =  \int Q(\bs{z}|\bs{x}) \log P(\bs{X}|\bs{z}) \diff \bs{z} 
   +\int Q(\bs{z}|\bs{x}) \log \frac{Q(\bs{z}|\bs{x})}{P(\bs{z}|\bs{x})} \diff \bs{z} \\
& - \int Q(\bs{z}|\bs{x}) \log \frac{Q(\bs{z}|\bs{x})}{P(\bs{z})} \diff \bs{z} \\
   & =  E_{z\sim Q}[\log P(\bs{x}|\bs{z})]
     +\mathit{KL}\left(Q(\bs{z}|\bs{x}) \|P(\bs{z}|\bs{x})\right) \\
& - \mathit{KL}\left(Q(\bs{z}|\bs{x})  \| P(\bs{z})\right) \, .
\end{align*}
By rearranging, we obtain
\begin{equation}
\begin{split} \label{eq:elbo}
\log P(\mathbf{x}) &- \mathit{KL}\left[Q(\bs{z}|\bs{x})
\|P(\bs{z}|\bs{x})\right] \\
& = E_{z\sim Q}[\log P(\bs{x}|\bs{z})] - \mathit{KL}\left[Q(\bs{z}|\bs{x})  \| P(\bs{z})\right] \, .
\end{split}
\end{equation}
This equation is crucial to the development variational autoencoders. The
left-hand side represents the term to maximize, plus an error term due
to approximating the true posterior $P(\bs{z}|\bs{x})$ with
$Q(\bs{z}|\bs{x})$. A  good choice of $Q$ gives a small
(hopefully zero) approximation error, and
consequently allows to directly optimize the likelihood. The
right-hand side of the equation (called Evidence Lower Bound, ELBO)
represents the equivalent of the 
left-hand side, but with the added bonus that, for a given choice of
$Q$ it becomes tractable and amenable to optimization. In particular,
by noticing that 
\begin{equation*}
\begin{split}
\mathit{KL}&\left[\mathcal{N}(\bs{\mu}, \bs{\Sigma}) \|
  \mathcal{N}(\bs{m}, \bs{S})\right]=\\
  & \frac{1}{2}\left(log\frac{|\bs{S}|}{|\bs{\Sigma}|} -K 
+ \mathit{tr}\left(\bs{S}^{-1}\bs{\Sigma}\right) 
+ \left(\bs{m} - \bs{\mu}\right)^T\bs{S}^{-1}\left(\bs{m} -
  \bs{\mu}\right)
\right) \, ,
\end{split}
\end{equation*}
we can choose to model  $Q(\bs{z}|\bs{x}) = \mathcal{N}(\bs{z}|\mu_\lambda(\bs{x}),
\Sigma_\lambda(\bs{x}))$, so that the term $\mathit{KL}\left[Q(\bs{z}|\bs{x})  \|
  P(\bs{z})\right]$ has a closed form.  
Again, we can model the parameters $\mu_\lambda(\bs{x})$
and $\Sigma_\lambda(\bs{x})$ through a neural
network trained on $\bs{x}$ and parameterized by $\lambda$. A
particular case that we shall use throughout this paper is given by 
$\bs{\Sigma}_\lambda (\bs{x}) = \mathit{diag}\left(\sigma_{\lambda,1}
  (\bs{x}), \ldots, \sigma_{\lambda,K} (\bs{x})\right)$. 

As a consequence, the modeling can resort to two neural networks: the
first one for ``encoding'' an input $\bs{x}$ into a latent variable
$\bs{z}$ by means of $Q_\lambda(\bs{z} |\bs{x})$. The second one
``decodes'' the latent representation $\bs{z}$ into the corresponding
$\bs{x}$ by means of $P_\phi(\bs{x}|\bs{z})$. The learning process is
governed by the loss function given by the right-hand side of
eq.~\ref{eq:elbo}, computed on the training data \bs{X}.

A problem with this loss is that term $E_{\bs{z}\sim Q}[\log
P(\bs{x}|\bs{z})]$ is typically approximated by 
sampling the values $\bs{z}$ according to $Q(\bs{z} |\bs{x})$. However, the 
sampling is a nondeterministic function that depends on the parameters
$\bs{\mu}_\lambda$ and $\bs{\Sigma}_\lambda$ (which in turn depend on $\lambda$) and
is not differentiable. To overcome 
this, \cite{kingma14} propose a \emph{reparametrization trick}: instead
of sampling \bs{z}, we can sample an auxiliary noise variable
\bs{\epsilon} according to a fixed distribution $P(\bs{\epsilon})$,
and obtain \bs{z} by means of a differentiable transformation
depending on $\lambda, \bs{\epsilon}$ and $\bs{x}$. Specifically, we can sample
gaussian noise $\bs{\epsilon} \sim \mathcal{N}(\bs{0}, \bs{I})$, and
obtain $\bs{z}_\lambda(\bs{\epsilon},\bs{x}_i) =
\bs{\mu}_\lambda(\bs{x}_i) +
\bs{\Sigma}_\lambda(\bs{x}_i)\cdot\bs{\epsilon}$,   normally 
distributed according to parameters $\bs{\mu}_\lambda$ and $\bs{\Sigma}_\lambda$. 

To summarize, given the loss function
\begin{equation}\label{eq:vaeloss}
\begin{split}
\mathcal{L}(\phi, \lambda; \bs{X})  =  \sum_i &\left\{  
\frac{1}{2}
    \sum_k \left(\sigma_{\lambda,k} (\bs{x}_i)  - 1 - 
      \log\sigma_{\lambda,k}(\bs{x}_i) 
    + \mu_{\lambda,k}(\bs{x}_i)^2
\right)\right.\\
& - \left.
\vphantom{\sum_{k} k} 
  E_{\bs{\epsilon} \sim \mathcal{N}(\bs{0}, \bs{I})} \left[
    \log P_\phi\left(\bs{x}_i | \bs{z}_\lambda(\bs{\epsilon}, \bs{x}_i)\right) \right]
\right\} \, ,
\end{split}
\end{equation}
we can learn the parameters $\lambda$ and $\phi$, and consequently
the encoder $Q_\lambda(\bs{z} | \bs{x})$ and the decoder
$P_\phi(\bs{x} | \bs{z})$.

\section{Variational Autoencoders for User Preferences}\label{sec:vae}

The general framework described in the previous section can be
instantiated to the case of collaborative filtering by specifying the
\bs{x} variables, and consequently the probability density
$P_\phi(\bs{x} | \bs{z})$. We analyze some alternative models here. We
shall use the following shared notation: $u \in U =\{1,
  \ldots, M\}$ indexes a user and $i \in I = \{1, \ldots, N\}$ indexes an
item for which the user can express a preference. We model implicit
feedback, thus assuming a preference matrix $\bs{X} \in
\{0,1\}^{N\times M}$, so that $\bs{x}_u$ represents the (binary) row
with all the item preferences for user $u$. Given $\bs{x}_u$, we define
$I_u= \{i\in I | x_{u,i} = 1 \}$ (with $N_u = |I_u|$). Analogously, $U_i= |\{u\in U |
x_{u,i} = 1 \}|$ and $M_i = |U_i|$. 

We also consider a precedence and temporal relationships
within \bs{X}. First of all, the preference matrix induces a natural ordering
relationship between items: $i \prec_u j$ has the meaning that
$x_{u,i} > x_{u,j}$ in the rating matrix.  Also, we assume the
existence of timing information $\bs{T}\in
{\rm I\!R}_{+}^{M\times N}\cup \{\emptyset\}$, where the term $t_{u,i}$
represents the time when $i$ was chosen by $u$ (with $t_{u,i} =
\emptyset$ if $x_{u,i} = 0$). Then, $i <_u j$ denotes that $t_{u,i} <
t_{u,j}$, With an abuse of notation, we also introduce a temporal mark
in the elements of $\bs{x}_u$: the term $\bs{x}_{u(t)}$ (with $1 \leq t
\leq N_u$) represents the $t$-th item in $I_u$ in the sorting induced
by $<_u$, whereas $\bs{x}_{u(1:t)}$ represents the sequence
$\bs{x}_{u(1)}, \ldots, \bs{x}_{u(t)}$. 

\subsection{Multinomial model}
\label{sec:mvae}

The reference model is the Multinomial variational autoencoder (MVAE)
proposed in \cite{Liang:2018:VAC:3178876.3186150}. Within this
framework, for a given user $u$ it is possible to devise $\bs{x}_u$
according to the generative setting:
\begin{equation}
\begin{gathered}
\bs{z}_u \sim \mathcal{N}(\bs{0}, \bs{I}_K) \, , \qquad  \pi(\bs{z}_u)
\sim \exp\left\{f_\phi(\bs{z}_u)\right\} \, , \\
 \bs{x}_u \sim \mathit{Multi}\left(N_u,\pi(\bs{z}_u)\right) \, .
\end{gathered}
\end{equation}
The underlying ``decoder'' is modeled by 
$$
\log P_\phi(\bs{x}_u |\bs{z}_u) = \sum_i x_{u,i} \log \pi_i(\bs{z}_u) \, , 
$$
thus enabling a complete specification of the overall variational
framework. Prediction for new items is accomplished by resorting to
the learned functions $f_\phi$ and $Q_\lambda$: given a user history
$\bs{x}$, we compute $\bs{z} = \bs{\mu}_\lambda(\bs{x})$ and then
devise the probabilities for the whole item set through
$\pi(\bs{z})$. Unseen items can then be ranked according to
their associated probabilities.

\subsection{Ranking Models}
\label{sec:bpr}

A different formulation is inspired by the Bayesian Personalized Ranking
(BPR) model introduced in
\cite{Rendle:2009}. Here, we explicitly model an
order among observations, so that for each triplet $u,i,j$ we can
devise whether $i \prec_u j$. In classical BPR, this is modeled
by associating a factorization rank $\bs{p}^T_u \bs{q}_i$ for each
pair $(u,i)$, and requiring that $\bs{p}^T_u \bs{q}_i < \bs{p}^T_u
\bs{q}_j$ whenever $i \prec_u j$. In a VAE setting, $\bs{p}^T_u
\bs{q}_i$ is replaced by a score based on the latent gaussian variable
$\bs{z}_{u,i}$: 
\begin{equation}
\begin{gathered}
\bs{z}_{u,i} \sim \mathcal{N}(\bs{0}, \bs{I}_K) \, ,\\
i \prec_u j \sim \mathit{Bernoulli}\left(\pi(\bs{z}_{u,i},
   \bs{z}_{u,j})\right) \, ,
\end{gathered}
\end{equation} 
Two different specifications for $\pi(\bs{z}_{u,i}, \bs{z}_{u,j})$ are
possible: either we let the neural network implicitly models the
comparison between $i$ and $j$
\begin{equation}\label{eq:rvaep}
\pi(\bs{z}_{u,i}, \bs{z}_{u,j})
\sim \sigma\left(f_\phi(\bs{z}_{u,i}, \bs{z}_{u,j})\right) \, ,
\end{equation}
or alternatively, we can assume that the network simply provides the
score upon which to compare: 
\begin{equation}\label{eq:rvaed}
\pi(\bs{z}_{u,i}, \bs{z}_{u,j})
\sim \sigma\left(f_\phi(\bs{z}_{u,j}) -  f_\phi(\bs{z}_{u,i})\right) \, .
\end{equation}
Here, $\sigma$ represents the standard logistic function $\sigma(z) =
1/(1+\exp\{-z\})$.

These alternatives represent two different instantiations of the
Ranking Variational Autoencoder. The
difference between them is substantial at prediction time: given $u$
and a set $J \subset I$ of unseen items, in both models we need to
devise $\bs{z}_{u,i} = \bs{\mu}_\lambda(\bs{x}_u | x_{u,i} = 1)$ for
each $i\in J$. However, the instance of eq. \ref{eq:rvaep} requires to
organize the data as triplets $u,i,j$  and to 
compute $P_\phi(i \prec_u j)$ for each triplet. These probabilities
can be then exploited to devise an order among the items in
$U$. Inconsistencies can arise in this setting, which can in principle
prevent to produce a global rank order. 

Conversely, the instance of eq. \ref{eq:rvaed} can
directly sort items according to $f_\phi(\bs{z}_{u,i})$ and does not
produce inconsistencies, thus being better suited for ranking
items. In the following we shall only focus on this 
instance, that we shall refer to as RVAE.

\subsection{Sequential Models}\label{seq:sequential}

\begin{figure*}[th!]
 \centering
    \begin{subfigure}[t]{0.3\textwidth}
        \centering
          \includegraphics[height=1.2in]{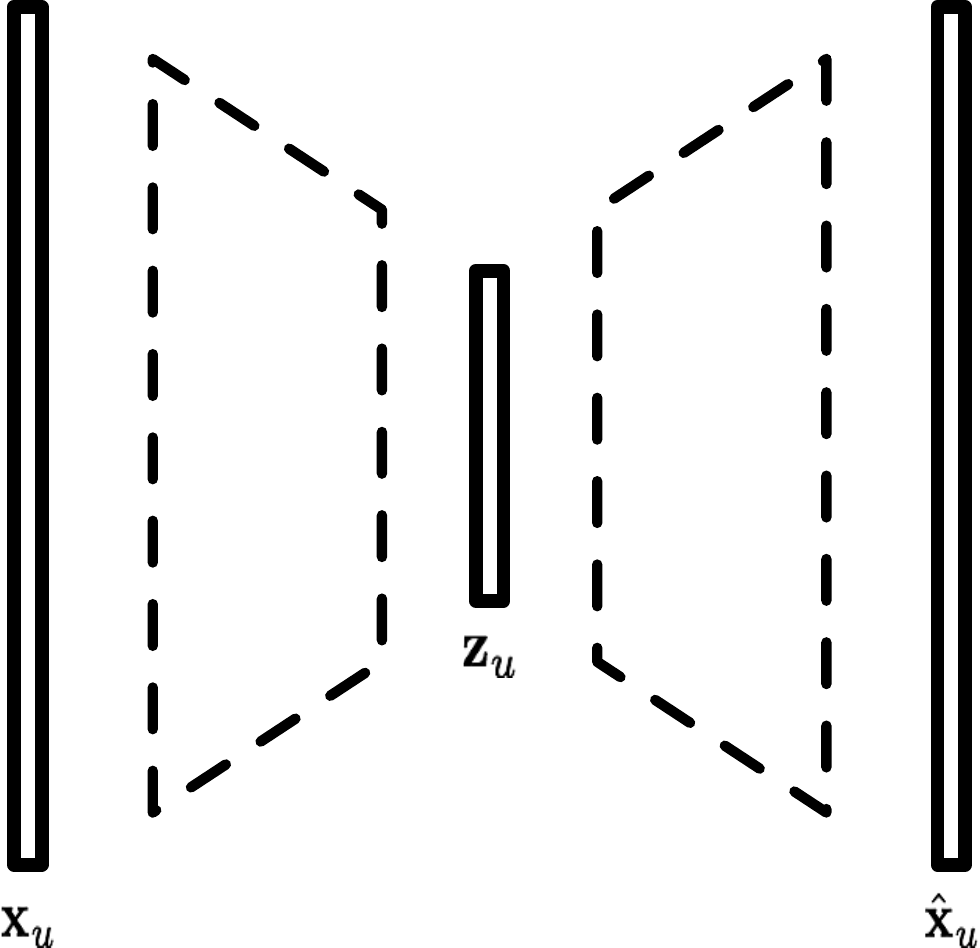}
        \caption{MVAE}
        \label{fig:mvae}
    \end{subfigure}
    ~ 
    \begin{subfigure}[t]{0.3\textwidth}
        \centering
          \includegraphics[height=1.2in]{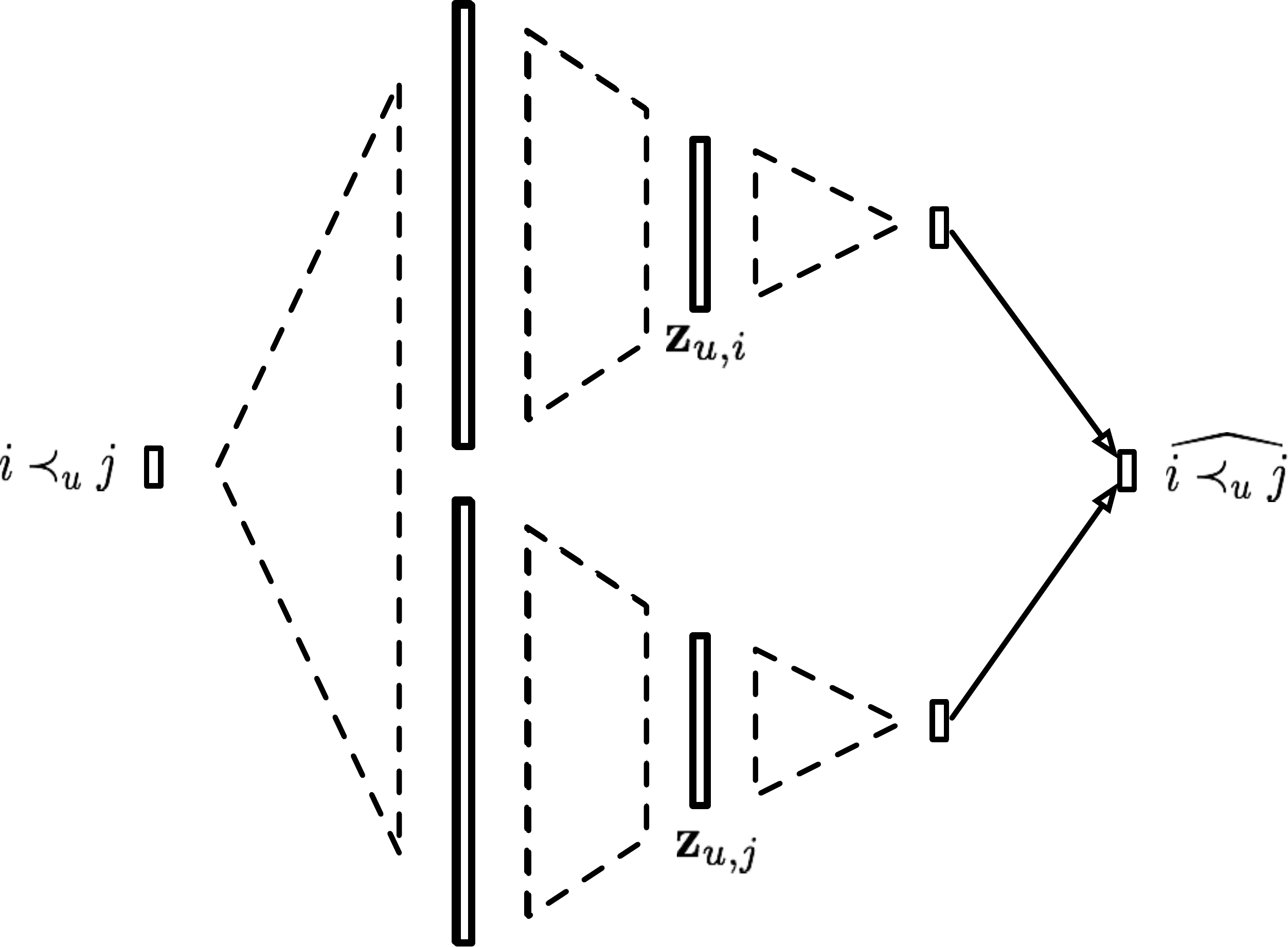}
        \caption{RVAE}
        \label{fig:rvae}
    \end{subfigure}
   ~ 
    \begin{subfigure}[t]{0.3\textwidth}
        \centering
          \includegraphics[height=1.2in]{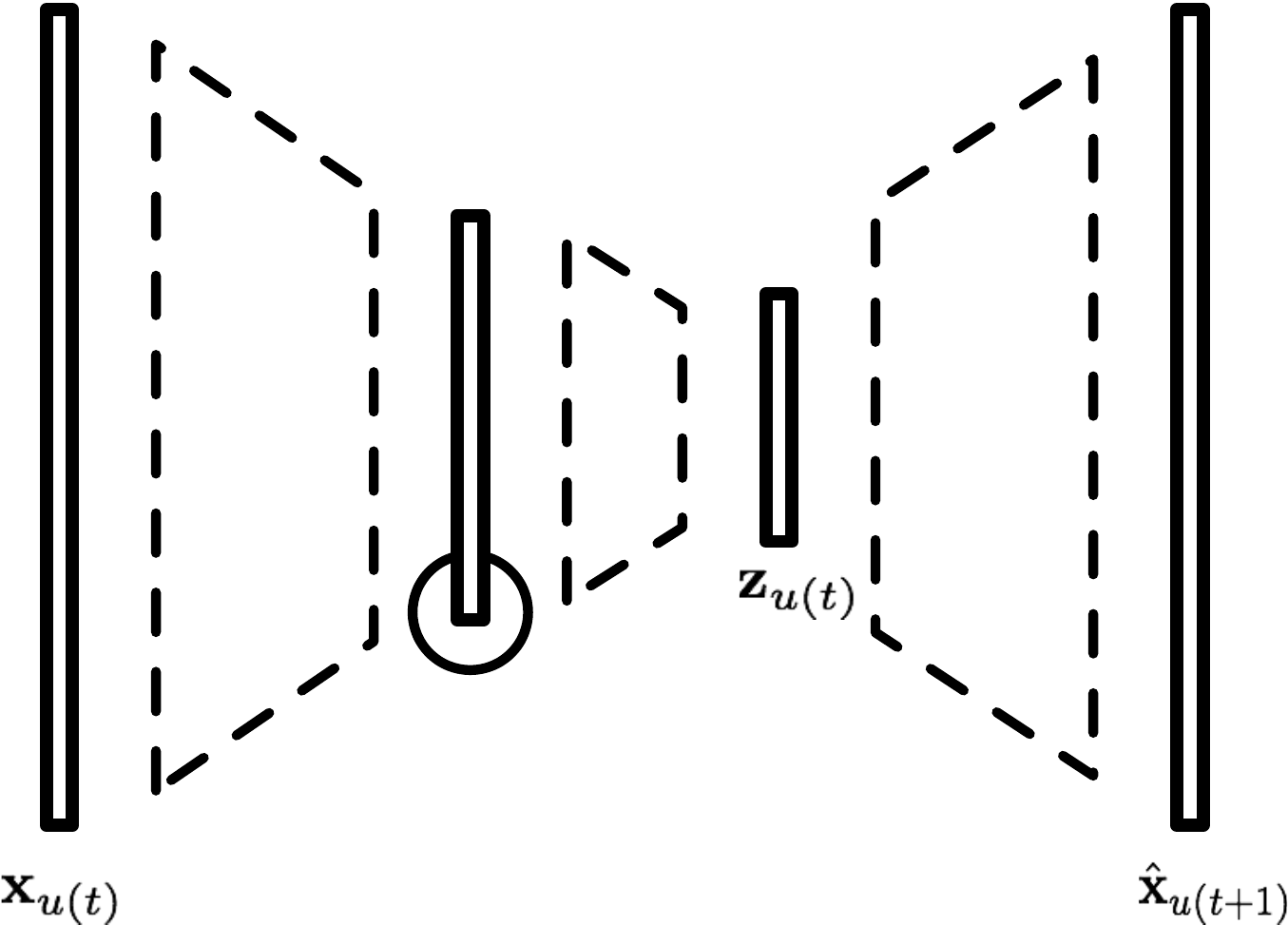}
        \caption{SVAE}
        \label{fig:svae}
    \end{subfigure}
    \caption{Variational architectures. Terms with 
    hat represent decoding reconstruction. Dotted boxes represent neural
    layers. }
    \label{fig:architecture}
\end{figure*}











The basic framework proposed in section \ref{sec:back} can also
be exploited to model time-aware user preferences. Ideally, latent
variable modeling should be able to express temporal dynamics and hence
causalities and dependencies among preferences in a user's
history. In the following we elaborate on this intuition.


\subsubsection{Modeling History-aware User Preferences}\label{sec:svae-simple}
Within a probabilistic framework, we can model temporal dependencies by
conditioning each event to the previous events: given a sequence
$\bs{x}_{(1:T)}$, we have
$$
P\left(\bs{x}_{(1:T)}\right) = \prod_{t=0}^{T-1}
P\left(\bs{x}_{(t+1)}| \bs{x}_{(1:t)}\right) \, .
$$
This specification suggests two key aspects: 
\begin{itemize}
\item There is a recurrent relationship between $\bs{x}_{(t+1)}$
  and $\bs{x}_{(1:t)}$ devised by
  $P\left(\bs{x}_{(t+1)}|\bs{x}_{(1:t)}\right)$, upon which the
  modeling can take advantage,   and 
\item each time-step can be handled separately, and in particular it
  can be modeled through a conditional VAE. 
\end{itemize}

Let us consider the following (simple) generative model 
\begin{equation}\label{eq:svae}
\begin{gathered}
\bs{z}_{u(t)} \sim \mathcal{N}\left(\bs{0},\bs{I}_K\right) \, , 
\qquad  \pi\left(\bs{z}_{u(t)}\right)
\sim \exp\left\{f_\phi\left(\bs{z}_{u(t)}\right)\right\} \, , \\[2mm]
 \bs{x}_{u(t)} \sim \mathit{Multi}\left(1,\pi\left(\bs{z}_{u(t)}\right)\right) \, ,
\end{gathered}
\end{equation}
which results in the joint likelihood
$$
P(\bs{x}_{u(1:T)}, \bs{z}_{u(1:T)}) = \prod_t
P(\bs{x}_{u(t)} |\bs{z}_{u(t)})P(\bs{z}_{u(t)}) \, .
$$
Here, we can approximate the posterior $P(\bs{z}_{u(1:T)}|
\bs{x}_{u(1:T)})$ with the factorized proposal distribution 
$$
Q_\lambda(\bs{z}_{u(1:T)}|\bs{x}_{(1:T)}) = \prod_t q_\lambda(\bs{z}_{u(t)}|\bs{x}_{(1:t-1)}) \, ,
$$
where $q_\lambda(\bs{z}_{u(t)}|\bs{x}_{(1:t-1)})$ is a gaussian
distribution whose parameters $\bs{\mu}_\lambda(t)$ and
$\bs{\sigma}_\lambda(t)$ depend upon the 
current history $\bs{x}_{u(1:t-1)}$, by means of a recurrent layer $\bs{h}_t$: 
\begin{equation}\label{eq:recurrent}
\begin{split}
\bs{\mu}_\lambda(t), \bs{\sigma}_\lambda(t) & =
                                                               \varphi_\lambda(\bs{h}_t)\\
\bs{h}_t & = \mathit{RNN}_\lambda\left(\bs{h}_{t-1},   \bs{x}_{u(t-1)} \right) \, .
\end{split}
\end{equation}

The resulting loss function follows directly from eq.~\ref{eq:vaeloss}: 
\begin{equation*}
\begin{split}
\mathcal{L}(\phi, \lambda,\bs{X})  =  \sum_u \sum_{t=1}^{N_u}&\left\{  
\frac{1}{2}
    \sum_k \left(\sigma_{\lambda,k} (t)  - 1 - 
      \log\sigma_{\lambda,k}(t) + \mu_{\lambda,k}(t)^2
\right)\right.\\
& - \left.
\vphantom{\sum_{k} k} 
  E_{\bs{\epsilon} \sim \mathcal{N}(\bs{0}, \bs{I})} \left[
    \log P_\phi\left(\bs{x}_{u(t)} | \bs{z}_\lambda(\bs{\epsilon}, t)\right) \right]
\right\} \, .
\end{split}
\end{equation*}

The proposal distribution introduces a
dependency of the latent variable from a recurrent layer, which allows
to recover the information from the previous history. We call this
model SVAE. Figure \ref{fig:architecture} shows the main
architectural difference with respect to the models proposed so
far. In SVAE, we can observe the recurrent relationship occurring in the layer
upon which $\bs{z}_{u(t)}$ depends. 

Notably, the prediction step can be easily accomplished in a similar
way as for MVAE: given a user history $\bs{x}_{u(1:t-1)}$, we can resort
to eq.~\ref{eq:recurrent} and set $\bs{z} =
\bs{\mu}_\lambda(t)$, upon which we can devise the probability for the
$\bs{x}_{u(t)}$ by means of $\pi(\bs{z})$.

\subsubsection{A taxonomy of sequential variational autoencoders}
\label{sec:svaetaxonomy}
We already discussed that within a sequence modeling framework, the
core of the approach is on modeling $P\left(\bs{x}_{(t+1)}|
  \bs{x}_{(1:t)}\right)$ through a conditional variational autoencoder. SVAE describes
just one of several possible modeling choices. 
\begin{figure*}[ht!]
 \centering
    \begin{subfigure}[t]{0.4\textwidth}
        \centering
        \begin{tabular}{c}
          \includegraphics[height=1.2in]{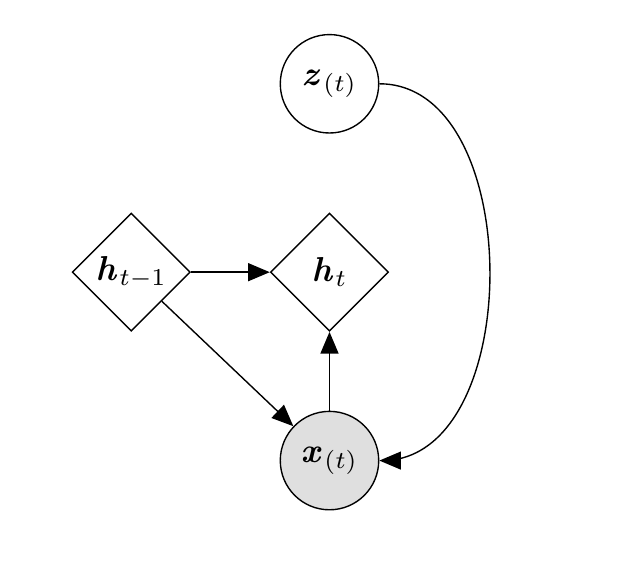}
          \\
          $            \scriptscriptstyle
            P\left(\bs{x}_{(1:T)}\right) = \int P\left(\bs{z}_{(1:T)}\right)\prod_{t=1}^{T-1} P_\phi\left(\bs{x}_{(t+1)}|
  \bs{z}_{(t+1)}, \bs{x}_{(1:t)}\right) \diff \bs{z}_{(1:T)}
          $
        \end{tabular}
        \caption{Single latent dependency}
        \label{fig:SLD}
    \end{subfigure}%
    ~ 
    \begin{subfigure}[t]{0.4\textwidth}
        \centering
        \begin{tabular}{c}
          \includegraphics[height=1.2in]{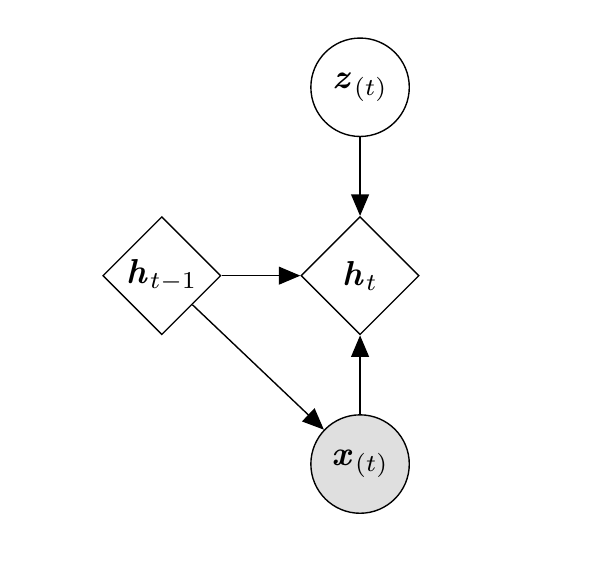}
          \\
          $
            \scriptscriptstyle
            P\left(\bs{x}_{(1:T)}\right) = \int P\left(\bs{z}_{(1:T)}\right)\prod_{t=1}^{T-1} P_\phi\left(\bs{x}_{(t+1)}|
              \bs{z}_{(1:t)},\bs{x}_{(1:t)}\right) \diff
            \bs{z}_{(1:T)}
          $
        \end{tabular}
        \caption{Multiple latent dependencies}
        \label{fig:MLD}
    \end{subfigure}
    \begin{subfigure}[t]{0.4\textwidth}
        \centering
        \begin{tabular}{c}
          \includegraphics[height=1.2in]{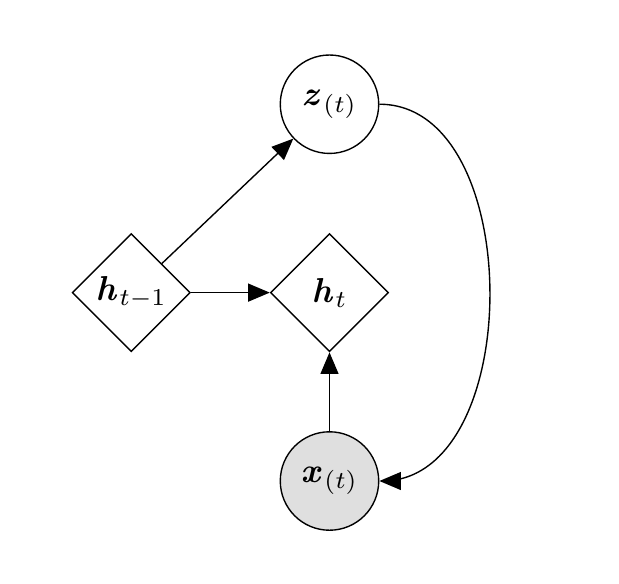}
          \\
          $
            \scriptscriptstyle
            P\left(\bs{x}_{(1:T)}\right) = \int\prod_{t=1}^{T-1} P_\phi\left(\bs{x}_{(t+1)}|
              \bs{z}_{(t+1)}\right)
            P\left(\bs{z}_{(t+1)}|\bs{x}_{(1:t)}\right)\diff
            \bs{z}_{(1:T)}
          $
        \end{tabular}
        \caption{Recurrent latent dependency}
        \label{fig:RLD}
    \end{subfigure}
    ~ 
    \begin{subfigure}[t]{0.4\textwidth}
        \centering
        \begin{tabular}{c}
          \includegraphics[height=1.2in]{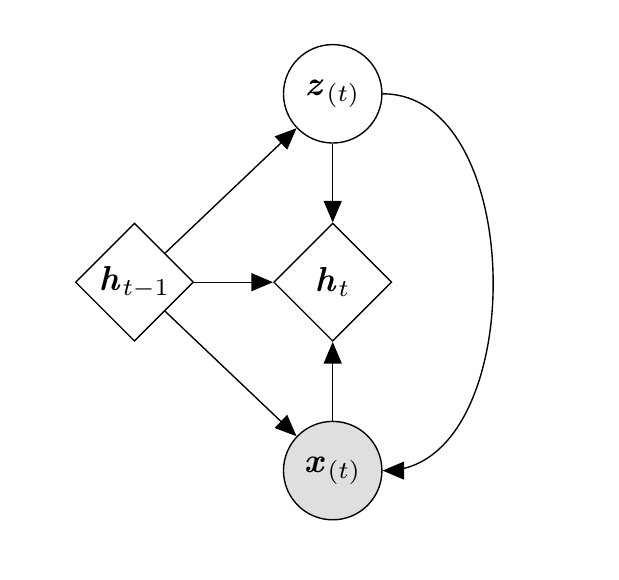}
          \\
          $
            \scriptscriptstyle
            P\left(\bs{x}_{(1:T)}\right) = \int\prod_{t=1}^{T-1} P_\phi\left(\bs{x}_{(t+1)}|
              \bs{z}_{(1:t+1)}, \bs{x}_{(1:t)}\right)
            P\left(\bs{z}_{(t+1)}|\bs{z}_{(1:t)}, \bs{x}_{(1:t)}\right)\diff
            \bs{z}_{(1:T)}
          $
        \end{tabular}
        \caption{Global recurrent dependency}
        \label{fig:GLD}
    \end{subfigure}
    \caption{Recurrent Variational Autoencoders. Diamond boxes
      represent deterministic variables.}
    \label{fig:recvae}
\end{figure*}
In fact, four alternate formalizations can take place, as illustrated in
fig.~\ref{fig:recvae}:  
\begin{itemize}
\item The simplest modeling, considers a single gaussian variable
  \bs{z} and a parameterization of the conditional distribution 
$P_\phi\left(\bs{x}_{(t+1)}| \bs{z},\bs{x}_{(1:t)}\right)$ with a function
$f_\phi(\bs{z}, \bs{h}_{t+1})$, where $\bs{h}_t$ represents the hidden
state of a recurrent neural function 
\begin{equation}\label{rnn-x}
  \bs{h}_t = \mathit{RNN}\left(\bs{h}_{t-1}, \bs{x}_{(t)}\right) \, .
\end{equation}
Figure \ref{fig:SLD} illustrates the graphical model and the sequence
likelihood.
\item Following \cite{BayerO14a}, we can alternatively introduce $t$
  independent gaussian variables and parameterize
  the conditional likelihood
$P_\phi\left(\bs{x}_{(t+1)}|
  \bs{z}_{(1:t)},\bs{x}_{(1:t)}\right)
$
by means of a function 
$f_\phi(\bs{h}_{t+1})$, where again $\bs{h}_t$ represents the hidden
state of a recurrent neural function 
\begin{equation}\label{rnn-x-z}
  \bs{h}_t = \mathit{RNN}\left(\bs{h}_{t-1}, \bs{x}_{(t)}, \bs{z}_{(t)}\right) \, .
\end{equation}
Figure \ref{fig:MLD} illustrates the graphical model and the sequence
likelihood. 
\item So far, the modeling combines history and latent variables to
  define the conditional distribution. An alternative consists in
  assuming that history affects the latent variable \bs{z} instead. In
  practice, the conditional 
  likelihood $P_\phi\left(\bs{x}_{(t+1)}| \bs{z}_{(t+1)}\right)$ would
  only depend on the latent variable, which exhibits a prior distribution
  $P\left(\bs{z}_{(t+1)}|\bs{x}_{(1:t)}\right)$, modeled as a gaussian
  with parameters depending on the current state $\bs{h}_t$ of the
  network, devised as in eq.~\ref{rnn-x}. The graphical model and
  sequence likelihood are shown in fig. \ref{fig:RLD} and it
  resembles the SVAE model discussed above. 
\item Finally, \cite{Chung:2015:RLV:2969442.2969572} propose a comprehensive
  model, where the gaussian latent variables $t+1$ also depend on each
  other through the Markovian dependency $P\left(\bs{z}_{(t+1)}|\bs{z}_{(1:t)},
    \bs{x}_{(1:t)}\right)$. Both this dependency and 
  $P_\phi\left(\bs{x}_{(t+1)}| \bs{z}_{(1:t+1)}, \bs{x}_{(1:t)}\right)$ can be specified by
  the hidden state $\bs{h}_{t}$ of a recurrent network, devised as in
  eq. \ref{rnn-x-z}.  The graphical model and
  sequence likelihood are shown in fig. \ref{fig:GLD}. 
\end{itemize}

\subsubsection{Extending SVAE}\label{sec:svae-ext}
The generative model of eq. \ref{eq:svae} only focuses on the
next item in the sequence. The base model however is flexible enough
to extend its focus on the next $k$ items, regardless of the time: 
$$
 \bs{x}_{u(t:t+k-1)} \sim \mathit{Multi}\left(k,\pi\left(\bs{z}_{u(t)}\right)\right),
$$
Again, the resulting joint likelihood can be modeled in different
ways. 
The simplest way consists in considering $\bs{x}$ as a time-ordered multi-set, 
\begin{equation}\label{eq:svae-ext-std}
  P(\bs{x}_{u(1:T)}, \bs{z}_{u(1:T)}) = \prod_t
  P(\bs{x}_{u(t:t+ k -1)} |\bs{z}_{u(t)})P(\bs{z}_{u(t)}) \, .
\end{equation}
Alternatively, we can consider the probability of an item as a mixture
relative to all the time-steps where it is considered:
\begin{equation}\label{eq:attention}
P(\bs{x}_{u(1:T)}, \bs{z}_{u(1:T)}) = \prod_t P(\bs{z}_{u(t)})
\sum_{t-k +1\leq j \leq t}P(\bs{x}_{u(t)} |\bs{z}_{u(j)}), 
\end{equation}
where $P(\bs{x}_{u(t)} |\bs{z}_{u(j)})$ is the probability of
observing $\bs{x}_{u(t)}$ according to $\pi(\bs{z}_{u(j)})$. 
In both cases, the variational approximation is modeled exactly as
shown above, and the only difference lies in the second component of
the loss function, which has to be adapted according to the above
equations.  


There is an interesting analogy between eq. \ref{eq:attention} and the
attention mechanism \cite{Vaswani:2017}. In fact, it can be noticed in
the equation that the prediction of $\bs{x}_{u(t)}$ depends on the
latent status of the $k$ previous steps in the sequence. In practice,
this enables to capture short-term dependencies and to encapsulate
them in the same probabilistic framework by weighting the likelihood
based on $\bs{z}_{u(t-k+1)}, \ldots, \bs{z}_{u(t)}$.

\section{Evaluation}\label{sec:exp}

We evaluate SVAE on some benchmark datasets, by comparing with various
baselines and the current state-of-the-art competitors, in order to
assess its capabilities in modeling preference data. Additionally, we
provide a sensitivity analysis relative to the configurations/contour
conditions upon which SVAE is better suited. The main highlight from
our experiments is that SVAE provides a huge edge over the current
state-of-the-art for the task of top-N recommendation across various
metrics.

\vspace*{-5mm}
\subsection{Datasets}
\begin{figure}[h!]
\begin{minipage}[t]{0.45\linewidth}
    \includegraphics[height=1.3in]{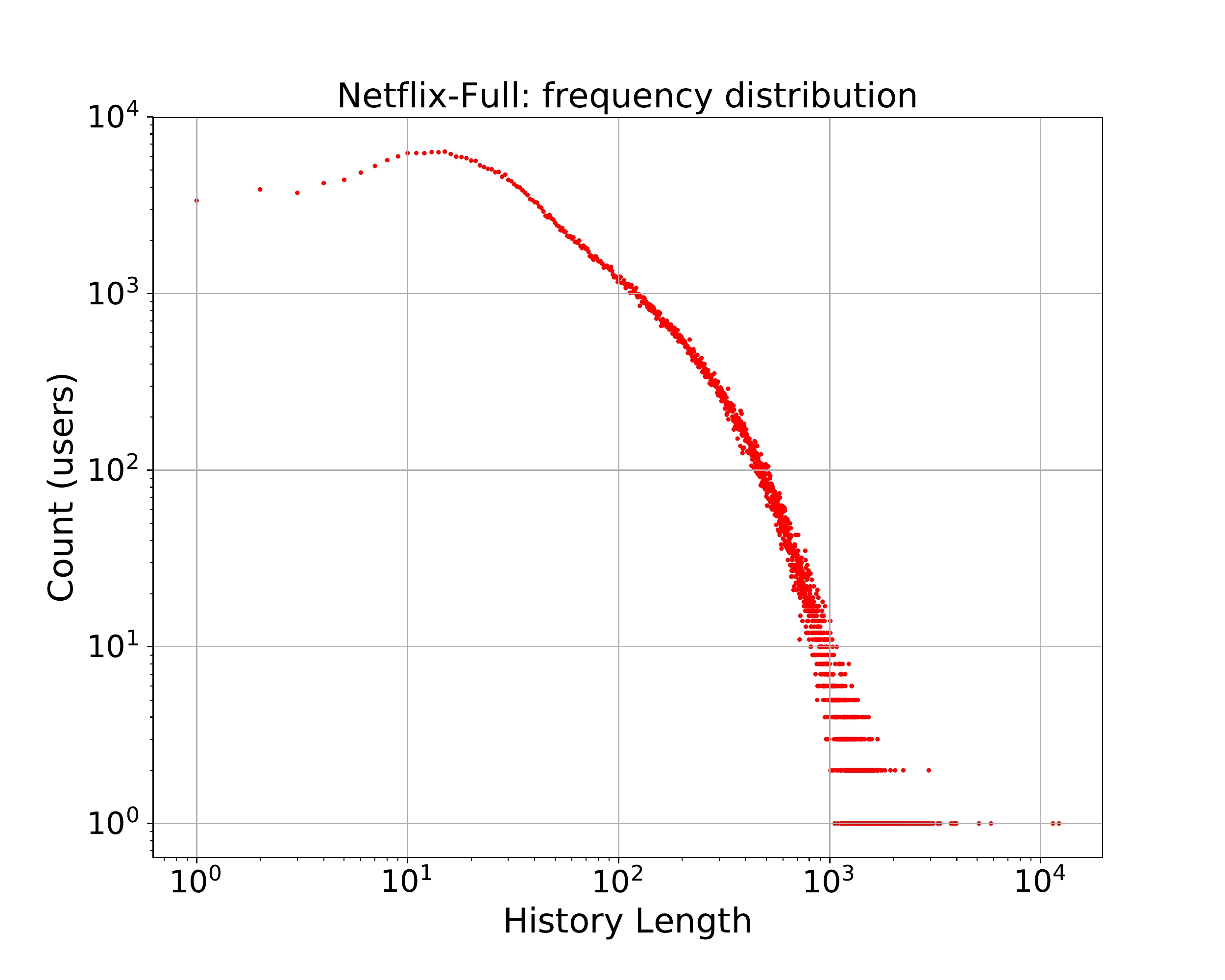}
\end{minipage}%
    \hfill%
\begin{minipage}[t]{0.45\linewidth}
    \includegraphics[height=1.3in]{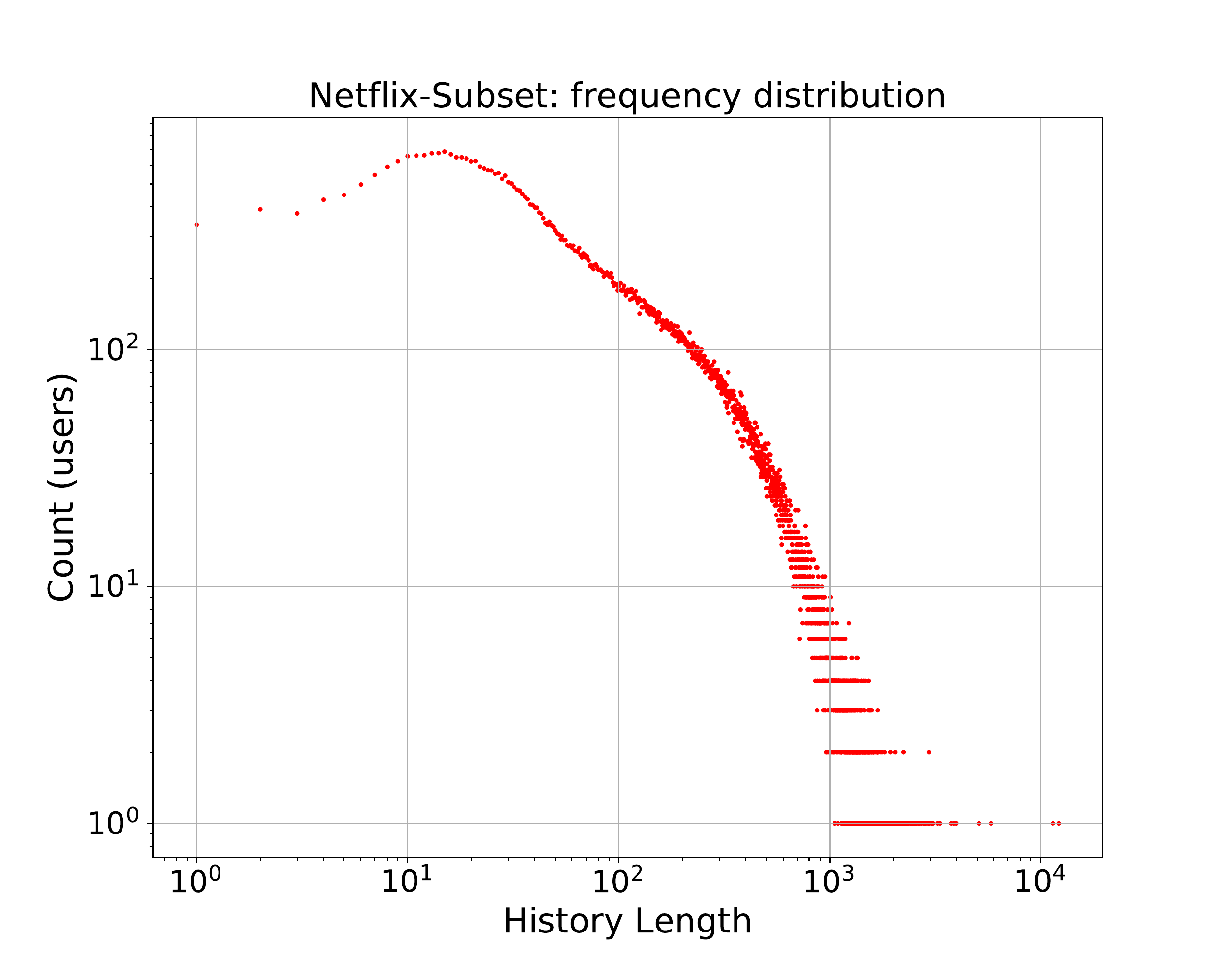}
  \end{minipage}
  \vspace{-5mm}
  \caption{Comparison of the user distributions of the full Netflix
    dataset and the subsample produced by progressive stratified
    sampling.}
    \label{fig:netflix_dist}
\end{figure}

We evaluate our model along with the competitors on two popular
publicly available datasets, namely \textit{Movielens-1M} and
\textit{Netflix}. 
\textrm{Movielens-1M} is a time series dataset containing
user-item ratings pairs along with the corresponding
timestamp. Since we work on implicit feedback, we binarize the
data, by considering only the user-item pairs where the rating provided
by the user was strictly greater than 3 on a range 1:5. 
\textrm{Netflix} has the same data format as \textrm{Movielens-1M} and
the same technique is used to binarize the ratings. We use a subset
of the full dataset that matches the user-distribution with the
full dataset. The subset is built by stratifying users according to
their history length, and then sampling a subset of size inversely
proportional to the size of the
strata. Figure~\ref{fig:netflix_dist} compares the distributions of
the full and the sampled dataset: We can notice that the
distributions share the same shape, but in the sample users with
small history length are undersampled whereas users with large
histories are kept.

\begin{table}
\begin{center}
  \includegraphics[width=0.4\textwidth]{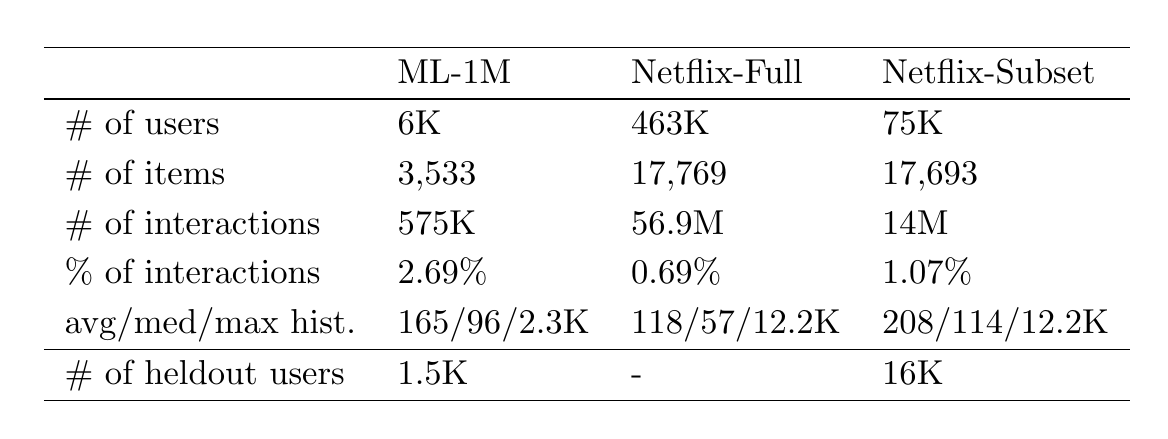}
\end{center}
\caption{Basic statistics of the datasets used in the experiments.}
\label{dataStats}
\end{table}

Table~\ref{dataStats} shows the basic statistics of the data. For
illustration purposes, we also show the basic statistics of the full
Netflix dataset. We can see that the average length of sequences in
the Netflix subset is significantly increased, as a result of
downsampling users with small history length. Also, notice that the
sampling procedure does not affect the number of items. 
To preprocess the data, we
first group the interacted items for each user, and ignore the users
who have interacted with less than five items. After preprocessing, we
split the remaining users into train, validation and test sets. 

\vspace*{-5mm}
\subsection{Evaluation Metrics and protocol}
Since we are considering implicit preferences, the evaluation is
done on top-n recommendation, and it relies on the following
metrics.
\begin{description}
\item[Normalized Discounted Cumulative Gain.] Also abbreviated as
  \textit{NDCG@n}, the metric gives more weight to the relevance of
  items on top of the recommender list and is defined as
$$
  \textit{NDCG@n} = \frac{\textit{DCG@n}}{\textit{IDCG@n}}
$$
where
$$
\textit{DCG@n} =\sum_{i=1}^{n} \frac{r_{i}}{log_{2}(i+1)}  \quad\text{and}\quad 
\textit{IDCG@n} = \sum_{i=1}^{|\textit{R}|} \frac{1}{log_{2}(i+1)}.
$$
Here, $r_i$ is the relevance (either 1 or 0 within the implicit
feedback scenario) of the $i$-th recommended items in the
recommendation list, and $R$ is the set of relevant items. 
\item[Precision.] By defining $\textit{Hits} = \sum_i r_i$ as the number of items
  occurring in the recommendation list that were actually
  preferred by the user, we have
  $$
  \vspace*{-1mm}
  \textit{Precision}@n = \frac{\textit{Hits}@n}{n}
  $$
\item[Recall,] defined as the percentage of items actually preferred by the user
  that were present in the recommendation list: 
$$
Recall@n  = \frac{\textit{Hits}@n}{|R|}
$$
\end{description}
In our experiments, we use the above metrics with two values of $n$,
respectively $10$ and $100$. 

The evaluation protocol works as follows.
We partition users into training, validation and test set. The
model is trained using the full histories of the users in the training
set. During evaluation, for each user in the validation/test
set we split the \textbf{time-sorted} user history into two parts,
\textit{fold-in} and \textit{fold-out} splits. The \textit{fold-in}
split is used as a basis to learn the necessary representations and provide a
recommendation list which is then evaluated with the \textit{fold-out} split
of the user history using the metrics defined above. 
We believe that this strategy is more robust when compared to
other methodologies wherein the same user can be in both the training
as well as testing sets. Table~\ref{dataStats} shows the number of
heldout users for each datasets. 

It is worth noticing that Liang et. al
\cite{Liang:2018:VAC:3178876.3186150} follow a similar strategy but
with a major difference: they do not consider any sorting for user
histories. That is, for the validation/test users, the \textit{fold-in}
set doesn't precede the \textit{fold-out} with respect to time. By
contrast, we keep the
\textit{fold-in} set to be the first \textit{80\%} of the time-sorted
user history, and the last $20\%$ represents the \textit{fold-out} set. We
shall see in the following sections that this difference is substantial in the
evaluation. 

\subsection{Competitors}
We compare our model with various baselines and current
state-of-the-art models including recently published neural 
architectures and we now present a brief summary about our
competitors to provide a better understanding of these models. 

\begin{itemize}
\item \textbf{POP} is a simple baseline where users are recommended
  the most popular items in the training set. 

\item \textbf{BPR}, already mentioned in section \ref{sec:bpr}, is a
  state of the art model based on Matrix
  Factorization, which ranks items
  differently for each user \cite{Rendle:2009}.
  There is a subtle
  issue concerning BPR: by separating users on
  training/validation/test as discussed above, the latent
  representation of users in the validation/test is not
  meaningful. That, is, BPR is only capable of providing meaningful
  predictions for users that were already exploited in the training
  phase. To solve this, we extended the training set to include
  the partial history in \textit{fold-in} for each user in the
  validation/test. The evaluation still takes place on their
  corresponding \textit{fold-out} sets. 

\item \textbf{FPMC} \cite{Rendle:2010} is a model which
  clubs both Matrix Factorization and Markov Chains together using
  personalized transition graphs over underlying Markov
  chains.

\item \textbf{CASER} \cite{Tang:2018}, already discussed in section
  \ref{sec:related}, is a convolutional model that uses vertical and
  horizontal convolutional layers to embed a sequence of recent items
  thereby learning sequential patterns for next-item
  recommendation. The authors have shown that this model outperforms
  other approaches based on recurrent neural network modeling, such as
  GRU4Rec. We use the implementation provided by the authors and tune
  the network by keeping the number of horizontal filters to be 16,
  and vertical filters to be 4. 

\item \textbf{MVAE}, discussed in section \ref{sec:mvae}, from which
  the SVAE model draws heavily. We use the implementation provided by
  the authors, with the default hyperparameter settings.

\end{itemize}

We also include the \textbf{RVAE} model proposed in section
\ref{sec:bpr}, that we consider a baseline here. Notably, despite
being considered a simple extension of the BPR
model, RVAE relies on a neural layer for embedding users: as a
consequence,  it does a better job in ranking items for the users which the
model has never seen before, contrary to BPR. In practice, RVAE
upgrades BPR to session-based recommendations. 

\subsection{Training Details}
The experiments only consider the SVAE model illustrated in subsection
\ref{sec:svae-simple} and the extensions of subsection~\ref{sec:svae-ext}. We
reserve a more detailed analysis of the extensions discussed in
subsection~\ref{sec:svaetaxonomy} to future work. 
The model is trained end-to-end on the full histories of the training
users.
Model hyperparameters are set using
the evaluation metrics obtained on validation users.

The SVAE architecture includes an embedding layer of size 256, a recurrent layer 
realized as a GRU with 200 cells, and two encoding layers (of size 150
and 64) and finally two decoding layers (again, of size 64 and 150).
We set the number $K$ of latent factors for the
variational autoencoder to be 64. Adam \cite{KingmaB14} was used to
optimize the loss function coupled with a weight decay of $0.01$.
As for RVAE, the architecture includes user/item embedding layers (of
size 128), two encoding layers (size 100 and 64), and a final layer
that produces the score $f_\phi(\bs{z}_{u,i})$. 
Both SVAE and RVAE were implemented in PyTorch \cite{paszke2017automatic}
and trained on a single GTX 1080Ti GPU.
The source code is available
on GitHub\footnote{\url{https://github.com/noveens/svae_cf}.}. 

\subsection{Results}

\begin{table*}
\begin{center}
  \includegraphics[width=0.8\linewidth]{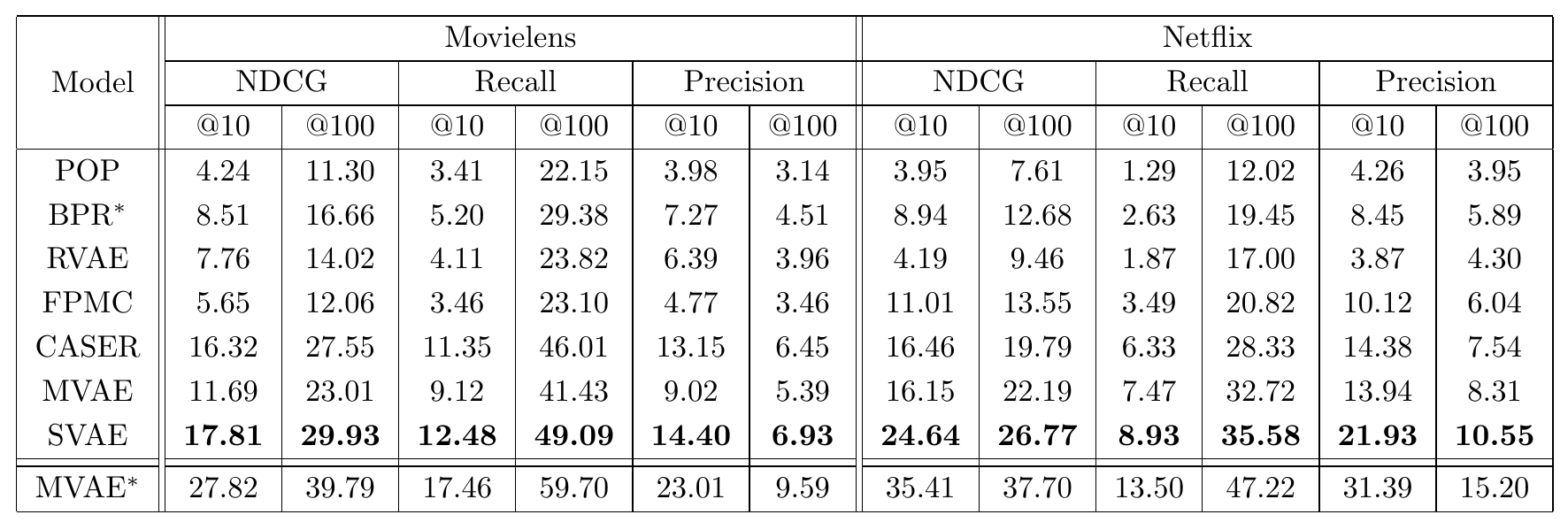}
\caption{Results of the evaluation (in percentage). MVAE$^\ast$ considers random
  splits that disregards the temporal order of user history.
  BPR$^\ast$ relies on including the \textit{fold-in} subsequences in
  the training phase.}
\label{table:resultsTable}
\end{center}
\end{table*}

In a first set of experiments, we compare SVAE with all competitors
described above. Table~\ref{table:resultsTable} shows the results of
the comparison. SVAE consistently outperforms the competitors on both
datasets with a significant gain on all the metrics. It is important
here to highlight how the temporal \textit{fold-in}/\textit{fold-out}
split is crucial for a fair evaluation of the predictive capabilities:
MVAE was evaluated both on temporal and random split, exhibiting
totally different performances. Our interpretation is that, with
random splits, the prediction for an item is easier if the encoding
phase is aware of forthcoming items in the same user history. This
severely affects the performance and overrates the predictive
capabilities of a model: In fact, the accuracy of MVAE drops
substantially when a temporal split is considered.

\begin{figure}[th!]
  \vspace*{-2mm}
  \centering
  \includegraphics[width=\columnwidth]{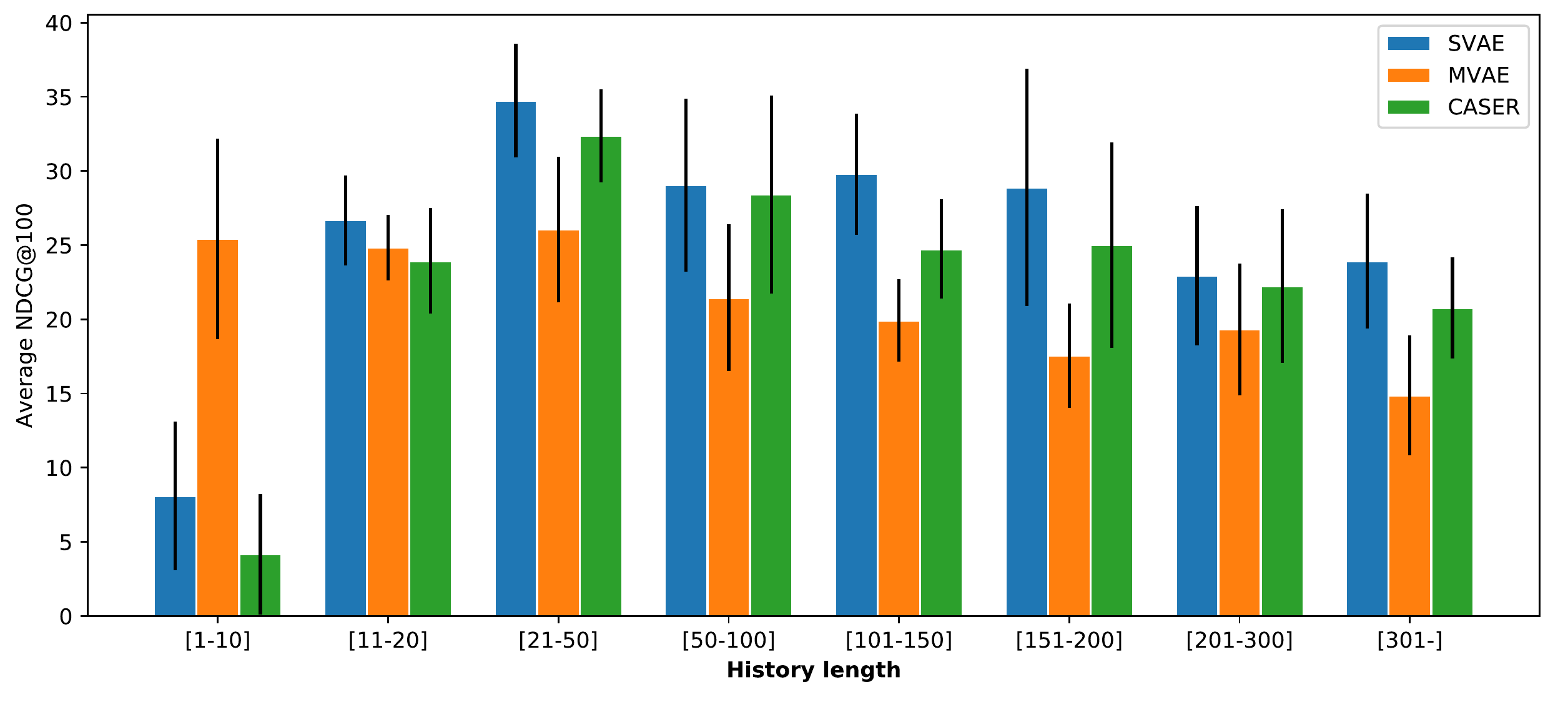}
  \vspace*{-8mm}
  \caption{Average NDCG@100 for MVAE, CASER \& SVAE across
    various history lengths.}
  \label{fig:seq_len_vs_ndcg}
\end{figure}

By contrast, SVAE is trained to capture the actual temporal
dependencies and ultimately results in better predictive accuracy.
This is also shown in fig.~\ref{fig:seq_len_vs_ndcg}, where we show
that SVAE consistently outperforms the competitors irrespective of the
size of \textit{fold-in}. The only exception is with very short
sequences (less than 10 items), where MVAE gets better results with
respect to the sequential models. It is also worth noticing how the
performance of both sequential models tend to degrade with increasing
sequences, but SVAE maintains its advantage over CASER. 

\begin{figure}[th!]
  \centering
  \includegraphics[height=1.2in]{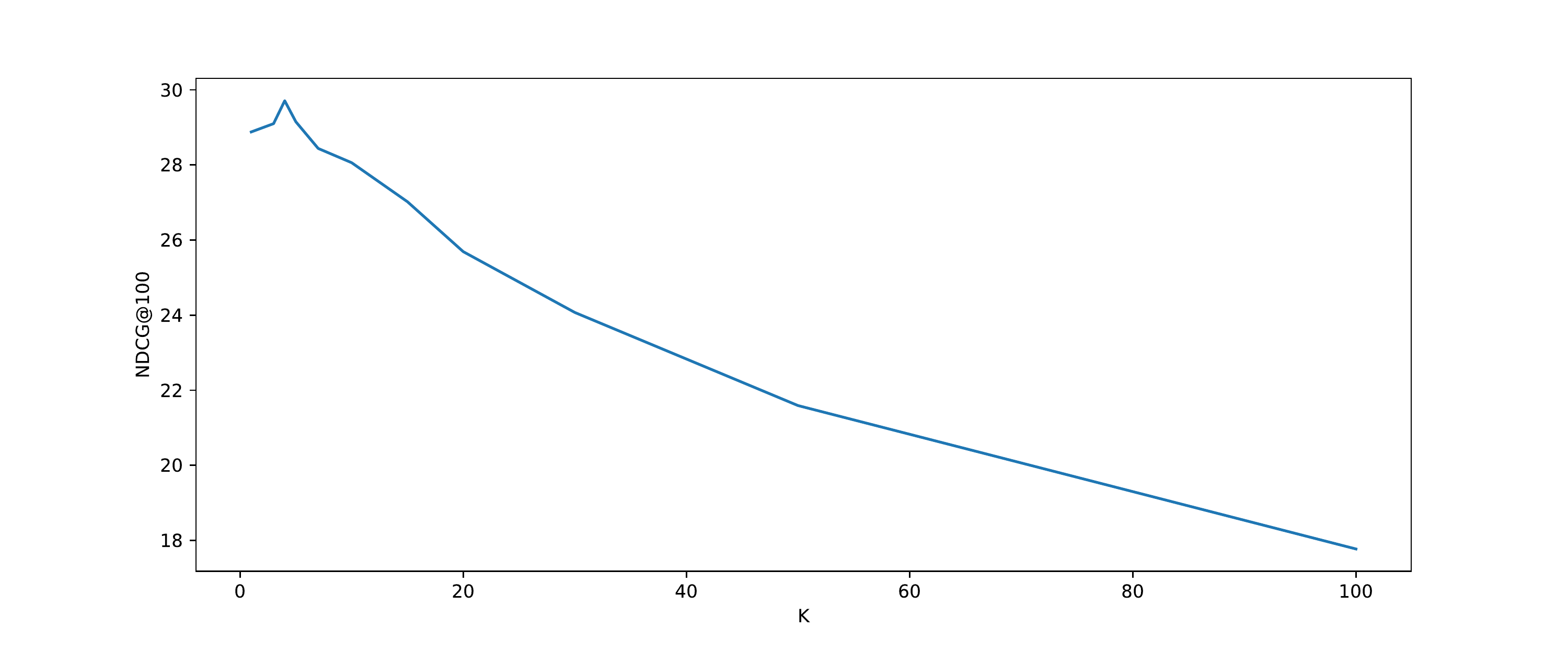}
        \vspace*{-4mm}
    \caption{NDCG@100 values for SVAE on Movielens, across different sizes for the
      number of items forward in time to predict on. 
}
    \label{fig:next_k_vs_ndcg}
\end{figure}

We discussed in section \ref{sec:svae-ext} how the basic SVAE framework can
be extended to focus on predicting the next $k$ items, rather then
just the next item. We analyse this capability in
fig.~\ref{fig:next_k_vs_ndcg}, where the accuracy for different values
of $k$ is considered according to the modeling in
\ref{eq:svae-ext-std}. On Movielens, the best value is achieved for
$k=4$, and acceptable values range within the interval $1-10$.

\begin{figure}[th!]
  \centering
  \includegraphics[height=1.4in,width=1.6in]{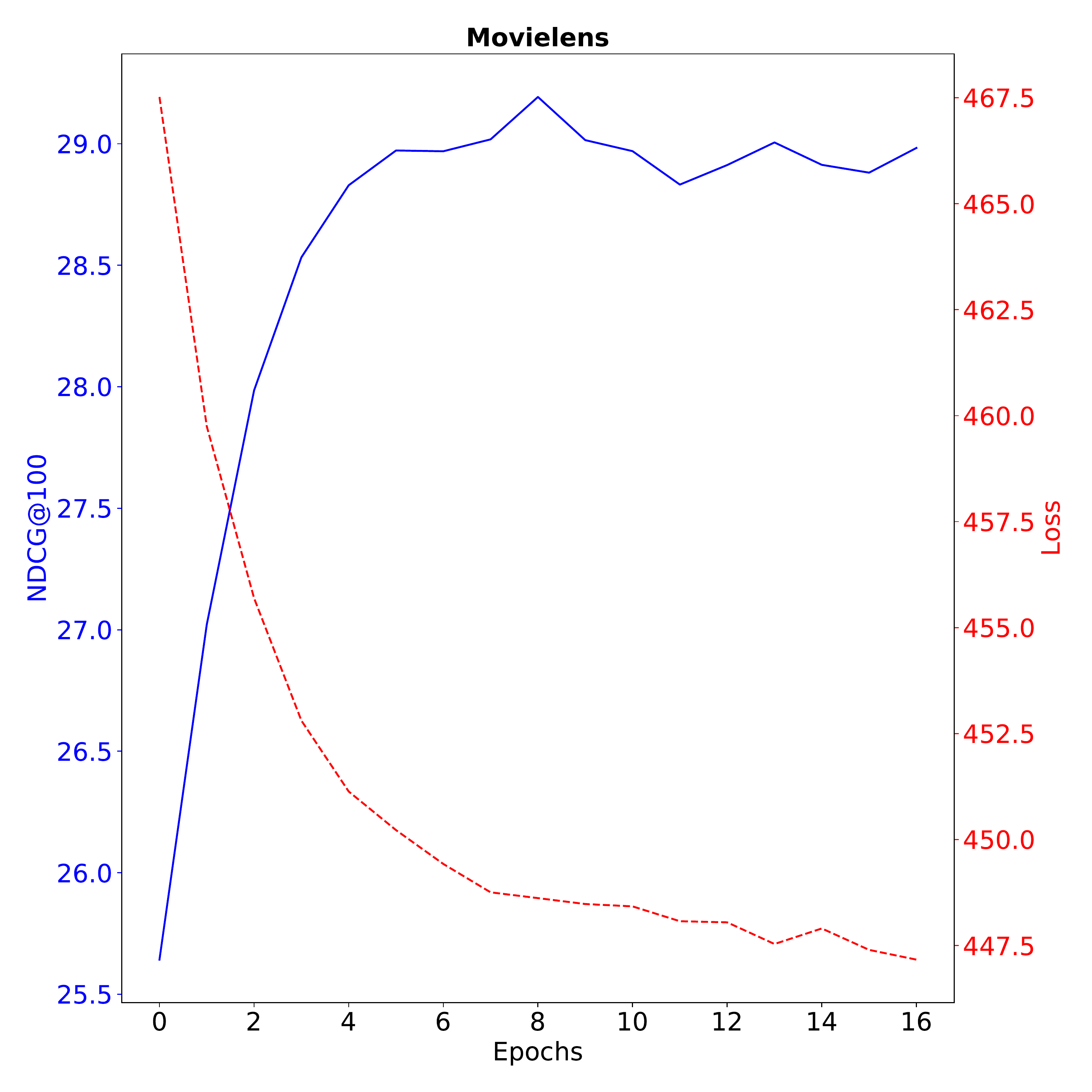}
  ~
  \includegraphics[height=1.4in,width=1.6in]{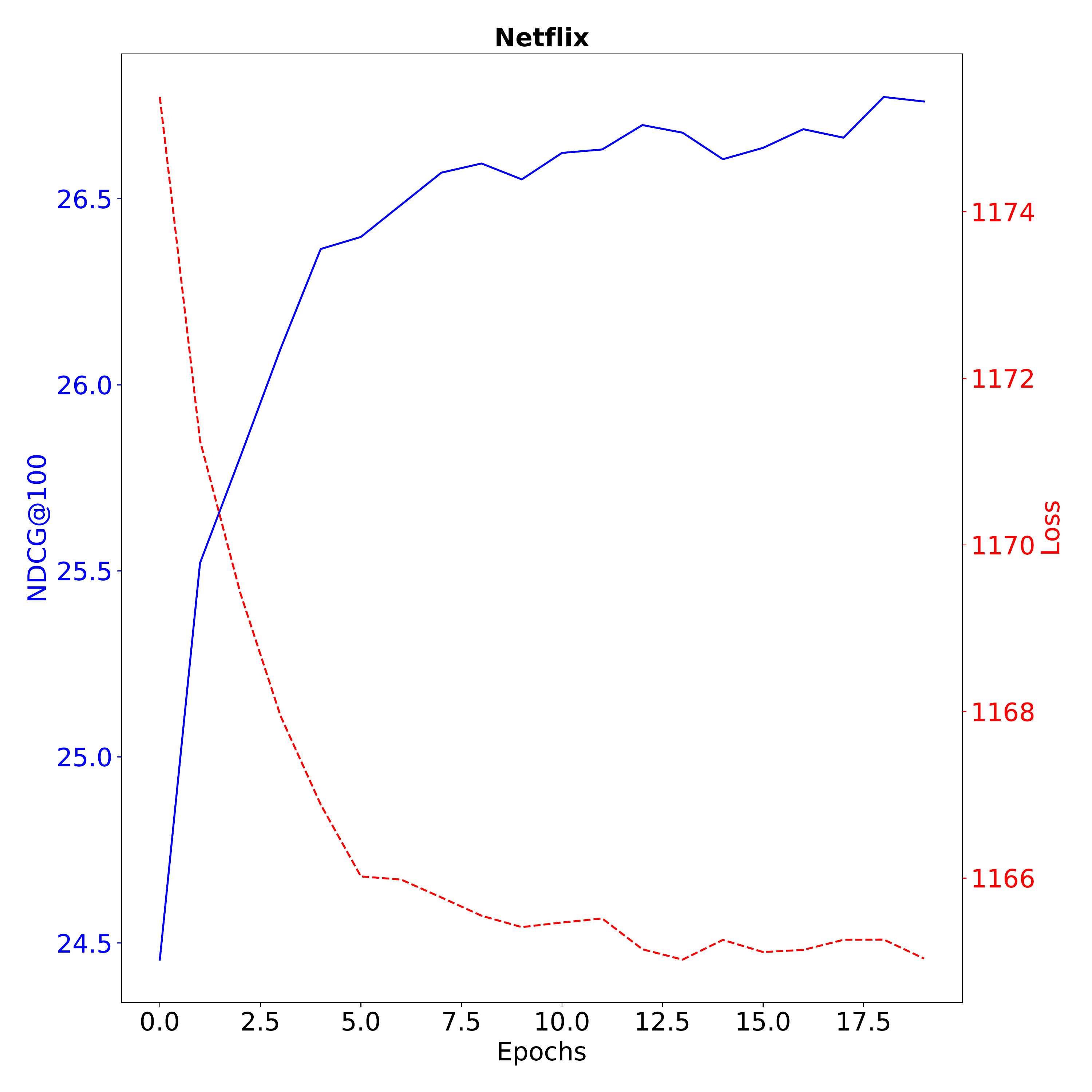}
  \vspace*{-4mm}
  \caption{Learning curves on validation data.}
  \label{fig:learningcurve}
\end{figure}

Finally, in fig.~\ref{fig:learningcurve} we analyse the convergence
rate of SVAE, in terms of NDCG (left y-axis and blue line) and loss
function values (right y-axis and red line). The learning phase
converges quickly and does not exhibit 
overfitting: on Movielens, a stable model is reached within 8
epochs, whereas Netflix requires 13 epochs. The average runtime per
epoch is 197 seconds on Movielens and 2 hours on Netflix: in practice,
the learning scales linearly with the number of
interactions in the dataset.

\section{Conclusions and future work}\label{sec:conc}

Combining the representation power of latent spaces, provided by
variational autoencoders, with the sequence modeling capabilities of
recurrent neural networks is an effective strategy to sequence
recommendation. To prove this, we devised SVAE, a simple yet robust
mathematical framework capable of modeling temporal dynamics 
upon different perspectives, within the fundamentals of variational
autoencoders. The experimental evaluation highlights the
capability of SVAE to consistently outperform state-of-the-art models.



The framework proposed here is worth further extensions
that we plan to accomplish in a future work. From a conceptual point
of view, we need to perform a thorough analysis of the taxonomy defined in section
\ref{sec:svaetaxonomy}. From an architectural point of view,
the attention mechanism, outlined in
section \ref{eq:attention}, requires a better understanding and a more
detailed analysis of its possible impact in view of the recent
developments \cite{Bahdanau14} in the literature. Also, the SVAE
framework relies on recurrent networks. However, different
architectures (e.g. based on convolution \cite{Tang:2018} or translation
invariance \cite{He:2017}) are worth being investigated within a
probabilistic variational setting.




\end{document}